\pdfoutput=1

\documentclass[final,5p,times,twocolumn,authoryear]{elsarticle}

\usepackage{amssymb}
\usepackage{listings}
\usepackage{booktabs} 
\usepackage{adjustbox}
\usepackage{url}
\usepackage{hyperref}
\usepackage{amsmath}

\newcommand\tabVSpace{0.2cm}
\newcommand\tabVSpaceDouble{0.4cm}

\journal{}

\begin{document}

\begin{frontmatter}



\title{DIAMBRA Arena: a New Reinforcement Learning Platform for Research and Experimentation}


\author{Alessandro Palmas - DIAMBRA - \texttt{alex@diambra.ai}}
\ead{alex@diambra.ai}

\begin{abstract}
The recent advances in reinforcement learning have led to effective methods able to obtain above human-level performances in very complex environments. However, once solved, these environments become less valuable, and new challenges with different or more complex scenarios are needed to support research advances. This work presents DIAMBRA Arena, a new platform for reinforcement learning research and experimentation, featuring a collection of high-quality environments exposing a Python API fully compliant with OpenAI Gym standard. They are episodic tasks with discrete actions and observations composed by raw pixels plus additional numerical values, all supporting both single player and two players mode, allowing to work on standard reinforcement learning, competitive multi-agent, human-agent competition, self-play, human-in-the-loop training and imitation learning. Software capabilities are demonstrated by successfully training multiple deep reinforcement learning agents with proximal policy optimization obtaining human-like behavior. Results confirm the utility of DIAMBRA Arena as a reinforcement learning research tool, providing environments designed to study some of the most challenging topics in the field.

\end{abstract}




\end{frontmatter}


\section{Introduction}
\label{introduction}

In the past few years, advances combining deep learing (DL) with reinforcement learning (RL) reached a few outstanding milestones \citep{Silver16, Silver17, Berner19}, showing that general algorithms can obtain top level performance even when facing complex tasks, while treating them as closed-boxes and without any problem-specific knowledge.

In order to compare different approaches, and to measure their rate of improvement, the research community needs a valid set of benchmarks. In RL the role of such benchmarks is played by software packages providing problems, also called \emph{environments}, that are challenging for RL to solve. In recent years, appearance of such tools relevantly contributed to the rapid development of this domain, making available problems of different complexity and high scalability.

Since simulation environments are the fundamental building block on which the RL community relays for testing its algorithms, their quality is of critical importance. In spite of that, the literature focused on this highly relevant aspect is not as comprehensive and developed as the one concerning algorithms and methods \citep{Juliani18}.

A large portion of current RL research platforms is based on popular video games or game engines such as Atari 2600, Quake III, Doom, and Minecraft. It has been a long time since artificial intelligence (AI) research started leveraging games to find meaningful, well posed and challenging problems, starting from chess and checkers \citep{Shannon50, Samuel59} and the application of RL to the game of Backgammon \citep{Tesauro95}. This trend continues today, as the AI research community shares the idea that creating algorithms able to reach human-level performances on these games, efficiently solving the same complex challenges, notably contributes to the pursue of machine intelligence \citep{Laird01}. 

As deep reinforcement learning (DeepRL) research finds new algorithms and above human-level performance is obtained, the benchmarks based on existing environments become less valuable and their impact as research drivers becomes marginal. There is a deep and entangled relation between algorithmic progress and new environments release, where the latter is key to favor the former. If on one side the research community is expected to advance the RL state-of-the-art developing better performing algorithms, who is in charge of delivering new high-quality environments is way less obvious. The task is in fact as challenging as algorithms development, requiring a relevant amount of time for design and implementation, as well as specialized domain knowledge. Proper care must be devoted to the creation of environments that pose meaningful challenges to learning systems, preferring properties that intercept main research interests to make them valuable benchmarks.

\begin{figure}[h]
\vskip 0.2in
\begin{center}
\centerline{\includegraphics[width=0.87\columnwidth]{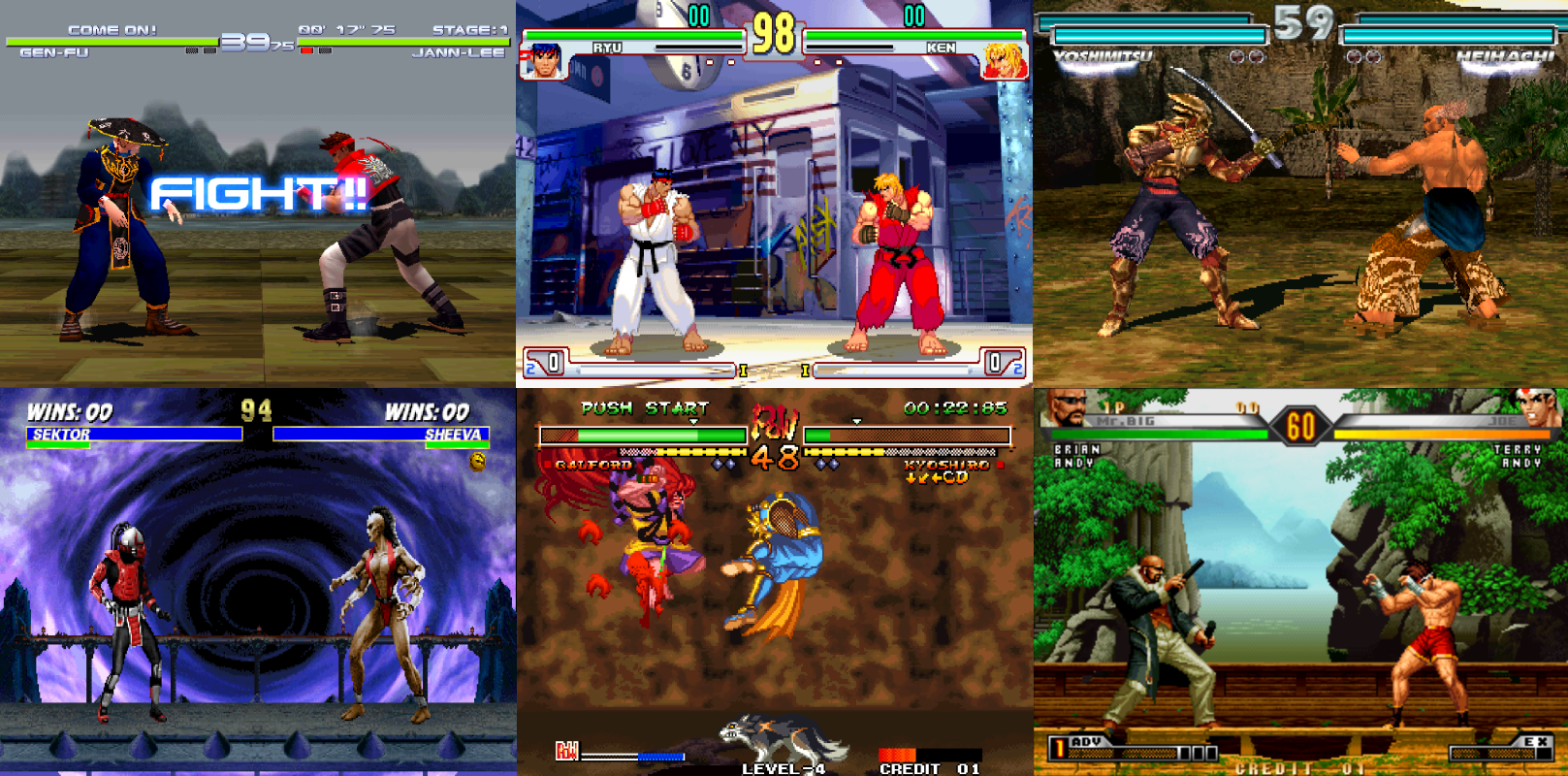}}
\caption{Snapshots of some DIAMBRA Arena environments}
\label{diambra-arena-envs}
\end{center}
\vskip -0.2in
\end{figure}

Within this context, this paper presents DIAMBRA Arena, a brand new software package featuring a collection of carefully selected, high-quality environments for RL research and experimentation. It has been developed with the goal of opening a high-quality curated stream of ever new environments, focusing on key aspects of RL research: competitive / cooperative multi-agents, agents transfer learning and generalization, hierarchical learning, human-agent cooperation / competition and human-in-the-loop training. 

It also aims at allowing everyone willing to approach RL to have access to state-of-the-art environments and to be able to use them. This means to design and develop something that is not extremely demanding in terms of hardware requirements, that runs on the majority of operating systems (OSs), well documented and provided with a comprehensive collection of working examples. 

Finally, it moves in the direction of applying on RL research, study and experimentation, the current \textit{gamification} trend, which is being successfully adopted in different fields of science lately \citep{Feger18, Feger19, Kalogiannakis21, Gari18, Hursen19}. Favoring exciting and rewarding experiences, this will be a key aspect in motivating and getting more people involved in the field.

The remainder of the paper is structured as follows: Section \ref{sec-background} discusses the state of the art of RL environments and provides an overview of relevant related work. Then, Section \ref{sec-arena}, presents DIAMBRA Arena software package and its most important technical details. In Section \ref{sec-agent} it is shown how DeepRL agents have been successfully trained in the games provided, presenting their architecture, the adopted training strategy and achieved performances and results. In Section \ref{sec-discussion}, one finds a discussion focusing on research directions that are enabled by DIAMBRA Arena, current limitations and the features roadmap planned for the near future. Finally, Section \ref{sec-conclusions} is where conclusions about this work are presented.

\section{Background and Related Work}
\label{sec-background}

RL problems, formulated in the standard form, assume there is an agent that is able to interact with an environment in iterative steps. Each iteration, the agent selects an action to execute, and it receives an observation and a reward from the environment. A RL algorithm aims at maximizing some measure of the agent total reward, as the agent interacts with the environment. In the RL literature, this formulation is typically formalized as a partially observable Markov decision process (POMDP). 

In the episodic declination of RL, the agent experience is split into a series of episodes. Each episode starts from an initial state and the interaction proceeds until a terminal state is reached. The goal in episodic RL is maximizing the expectation of total cumulative reward per episode, and to achieve a high level of performance in the lowest number of episodes possible. 

Around year 2012, the first benchmarks have started to arise in standard form, with the aim of creating a common test-bed to measure algorithms performances and to stimulate RL research. Since then, a few very notable environments and software packages have been developed and used to demonstrate tremendous improvements and outstanding breakthroughs in the RL domain, proving how important their role is in driving advancements in this field. Here is a list of the most relevant ones currently used in RL research, all publicly available:

\begin{itemize}
\item \textit{Arcade Learning Environment (ALE)} \citep{Bellemare12}: provides an interface to hundreds of Atari 2600 game environments with distinctive features. It has been and is still widely adopted as rigorous test-bed for evaluating and comparing approaches and algorithms in RL, representing the first important effort made in this standardization direction.
\item \textit{OpenAI Gym and OpenAI Retro} \citep{Brockman16, Nichol18}: OpenAI Gym includes a collection of benchmark environments that expose a common interface, providing different categories of problems like classic control and toy text, algorithmic, Atari and 2D and 3D robots. OpenAI Retro creates RL environments from various emulated video games.
\item \textit{DeepMind Lab} \citep{Beattie16}: first-person 3D game platform, it can be used to study autonomous artificial agents learning complex tasks in large, partially observed, and visually diverse worlds. Powered by a fast and widely recognized game engine, features a simple and flexible API that enables creative task-designs.
\item \textit{Starcraft II Learning Environment} \citep{Vinyals17}: based on the game StarCraft II, it represented a new grand challenge for RL, providing a more difficult class of problems than those considered in most previous research. Its main features are: multi-agent setting with multiple players interacting, imperfect information and partial observability, large action space, large state space, and delayed credit assignment requiring long-term strategies over thousands of steps.
\item \textit{Unity ML-Agents and Obstacle Tower} \citep{Juliani18, Juliani19}: Unity ML-Agents is open source project enabling to create simulated environments using the Unity Editor, and interact with them via a Python API. It includes a set of example environments together with state of the art RL, imitation learning and self-play algorithms in both symmetric and asymmetric games. Obstacle Tower is a high fidelity, 3D, 3rd person, procedurally generated environment. It provides both low-level control and high-level planning problems in tandem, learning from pixels and a sparse reward signal, as well as performance evaluation based on unseen instances of the environment.
\item \textit{VizDoom} \citep{Kempka16}:  based on the classical first-person shooter video game, Doom, uses raw visual information employing the first-person perspective in a semi-realistic 3D world. It is lightweight, fast, and highly customizable via a convenient mechanism of user scenarios.
\item \textit{MuJoCo} \citep{Todorov12}: physics engine focused on robotics, biomechanics, graphics and animation, and other areas where fast and accurate simulations are needed. Its key features are speed, accuracy and modeling power, specifically designed for model-based optimization, and in particular optimization through contacts. It has been one of the main references in the RL domain for robotics applications.
\end{itemize}

These environments are the most important actors in the field of RL platforms for research and experimentation. Section \ref{sec-arena} introduces DIAMBRA Arena, which aims at becoming a new player in this domain, providing new problems and challenges to help RL community advancing state-of-the-art. How it compares with those presented in this section in terms of features, speed and memory footprint is discussed in Section \ref{sec-rlenv-comparison}.

\section{DIAMBRA Arena}
\label{sec-arena}

This section presents DIAMBRA Arena software package and its most important technical details. It starts providing a high level overview, then it describes its technical details in terms of its basic usage, available settings, observation and action spaces, the reward wrappers and environment wrappers. Next subsection describes its advanced features, such as human-in-the-loop training and multi-agent and self-play, and the final subsection compares DIAMBRA Arena with other RL environments in terms of features and performances.

\subsection{Overview}

DIAMBRA Arena is a software package featuring a collection of high-quality environments for RL research and experimentation. It provides a standard interface to popular arcade emulated video games, offering a Python API fully compliant with OpenAI Gym format, that makes its adoption smooth and straightforward.

It is accompanied by a comprehensive documentation \citep{DiambraDocs} and its repository \citep{DiambraRepo} comes with a collection of examples covering main use cases of interest that can be run in just a few steps. It supports all major OSs (Linux, Windows and MacOS) and can be easily installed via Python PIP, as described in the installation section of the documentation. It is completely free to use, the user only needs to register on the official website.

The first version of the software focuses on fighting games, creating a robust and consistent package. All of them are episodic RL environments, with discrete actions that represent gamepad buttons. The observations to be used for control are composed by a screen pixels buffer plus additional data made of RAM values representing game elements such as characters health values or characters stage side. The problem can thus be framed as \textit{"control from pixels"} only, or as a more general one, taking advantage of RAM numerical values too. It is worth noticing that RAM values usage is guaranteed to be fair, meaning that they only provide redundant information that can be retrieved from the game screen alone, and no "hidden" state is contained in it.

Every environment supports both single player (1P) as well as two players (2P) mode. The former is the "classic" arcade race for clearing the game achieving the score record that can be used for \textit{standard RL}. The latter adds three new, orthogonal dimensions to these environments, making them suitable for research and experimentation in the domains of \textit{competitive multi-agent} as well as for \textit{human-agent cooperation / competition}, and allowing to make use of \textit{self-play} for training. 

In addition, it can be easily set up to explore \textit{human-in-the-loop training}, covering applications like \textit{human assisted rewards} and the natively supported \textit{imitation learning}, for which it provides tools to record human expert demonstrations and a specific class to seamlessly load and use them for agent training. 

All interfaced games have been selected among the most popular and successful fighting retro-games. They have been chosen so that, while sharing the same fundamental mechanics, they provide slightly different challenges, with specific features such as different number of characters to be used at the same time, how to handle combo moves, possibility to recharge health bars or not, and similar.

Whenever possible, games are released with all hidden/bonus characters unlocked. For every released title, extensive testing has been carried out, being the minimum requirement for a game to be released in \emph{beta}. After that, the next internal step is training a DeepRL agent to play, and possibly complete it, making sure the 1P mode is playable with no bugs up until game end. This is the condition under which titles are moved from \emph{beta} to \emph{stable} status.

\subsection{Technical Details}

\subsubsection{Basic Usage and Settings}
\label{basic-usage-settings}

\begin{figure}[h]
\vskip 0.2in
\begin{center}
\centerline{\includegraphics[width=0.85\columnwidth]{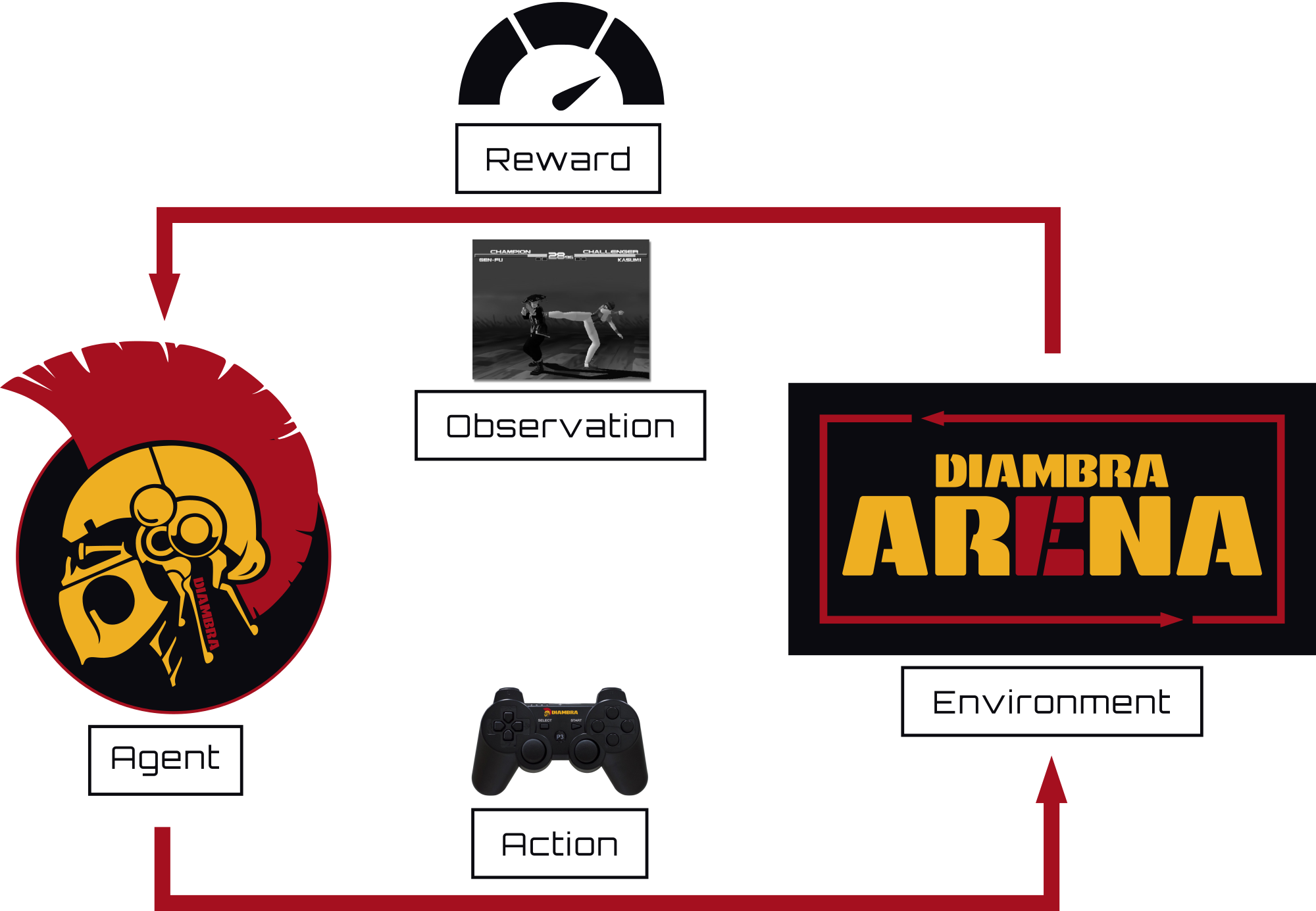}}
\caption{Environment interaction framed under the standard RL POMDP paradigm}
\label{basic-usage}
\end{center}
\vskip -0.2in
\end{figure}

DIAMBRA Arena usage follows the standard RL interaction framework: the agent sends an action to the environment, which processes it and performs a \emph{transition} accordingly, from the starting \emph{state} to the \emph{new state}, returning the \emph{observation} and the \emph{reward} to the agent to close the interaction loop. \textbf{Figure \ref{basic-usage}} shows this typical interaction scheme and data flow.

The shortest snippet for a complete basic execution of the environment consists of just 11 lines of code, and is presented in \textbf{Listing \ref{basic-usage-code}} below.

\lstset{frame=lines}
\lstset{label={basic-usage-code}}
\lstset{basicstyle=\footnotesize}
\begin{lstlisting}[language=Python, caption=DIAMBRA Arena basic usage, captionpos=b]
import diambra.arena                                                                                                                            
                                                                                                                                                      
env = diambra.arena.make("doapp")                                                                                                  
obs = env.reset()                                                       
                                                                                
while True:
    env.render()

    actions = env.action_space.sample()
    obs, rew, done, info = env.step(actions)

    if done:
        obs = env.reset()
        break

env.close()
\end{lstlisting}

A trained agent is expected to replace random action sampling with the result of its policy prediction based on the observation.

All environments share the list of options presented in \textbf{Table \ref{settings-table}}, allowing to handle many different aspects of them, controlled by key-value pairs in a Python dictionary. The high level presentation reported here is detailed in depth in the documentation. 

\begin{table*}[h]
\caption{Environment settings}
\begin{center}
\begin{small}
\begin{sc}
\begin{adjustbox}{max width=\textwidth}
\begin{tabular}{rl}
\toprule
Setting & Description \\
\midrule
Player & \begin{tabular}{@{}l@{}}Allows to select single player (1P) or two players (2P) mode. In 1P mode,\\
the same parameter is  used to select on which side to play, P1 (left) or P2 (right) \\\end{tabular}\\[0.2cm]
\midrule
Step ratio & 
\begin{tabular}{@{}l@{}}Defines how many steps the game (emulator) performs for every environment step.\\
This value has a direct impact on how frequently actions are sent to the game\end{tabular}\\[0.2cm]
\midrule
Frame shape & \begin{tabular}{@{}l@{}}Resizes game frame to prescribed height and width, keeping it  RGB or making it grayscale \end{tabular}\\
\midrule  
\begin{tabular}{@{}r@{}}Continue game\\
(1P mode only)\end{tabular}
& 
\begin{tabular}{@{}l@{}}Defines if and how to allow "Continue" when the agent is about to face the game over condition\end{tabular}\\
\midrule
\begin{tabular}{@{}r@{}}Difficulty\\
(1P mode only)\end{tabular} & Selects game difficulty\\[\tabVSpace]
\midrule
Character(s) &
Selects the character(s) the agent will use \\
\midrule
Characters outfit & 
\begin{tabular}{@{}l@{}}Defines the number of outfits to draw from at character selection\end{tabular}\\
\midrule
Action space & 
Specifies the action spaces to be used\\
\midrule
\begin{tabular}{@{}r@{}} Attack buttons\\
 combination \end{tabular}
 &  
\begin{tabular}{@{}l@{}} Specifies if to include attack buttons combinations or not as available actions \end{tabular} \\
\midrule
 Hardcore & 
\begin{tabular}{@{}l@{}}Limits the observation to the screen pixels discarding additional RAM states\end{tabular}\\

\bottomrule
\end{tabular}
\end{adjustbox}
\end{sc}
\end{small}
\end{center}
\vskip -0.1in
\label{settings-table}
\end{table*}

\subsubsection{Observation Space}
\label{obs-spaces-par}
Environment observations are composed by two main elements: a visual one (the game frame) and an aggregation of quantitative values called \textit{RAM states} (stage number, health values, etc.). In the default mode, both of them are exposed through an observation space of type \texttt{gym.spaces.Dict}\footnote{For details on Gym spaces, see \url{https://github.com/openai/gym/tree/master/gym/spaces/}}. 

\begin{figure}[h]
\vskip 0.2in
\begin{center}
\centerline{\includegraphics[width=\columnwidth]{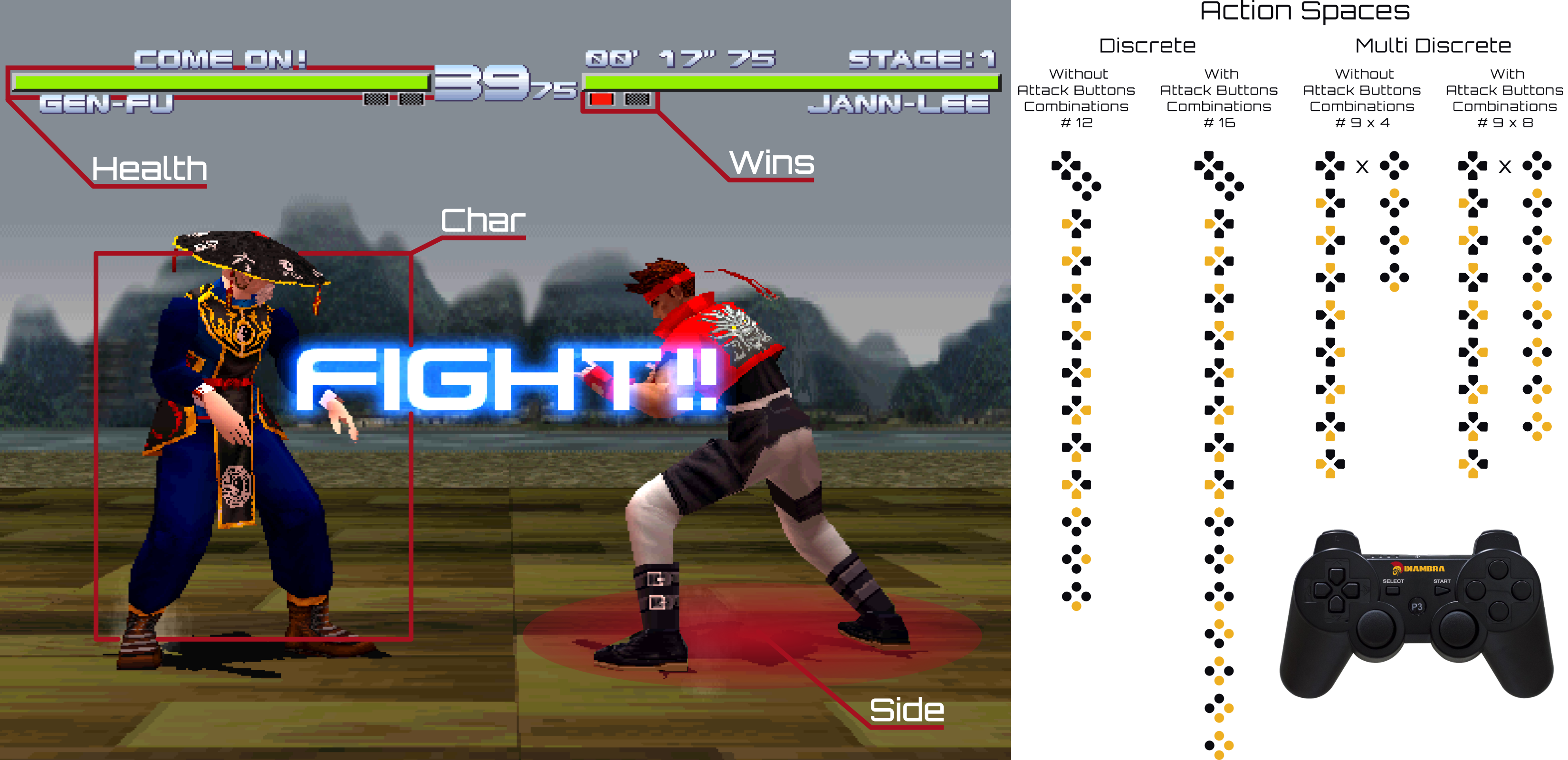}}
\caption{Representation of Dead Or Alive++ observation space (left) and action spaces (right). Observations are made by the game frame plus \textit{RAM states} (shown in the call-outs). The four action spaces are represented as sets of correspondent gamepad buttons. }
\label{observation-action-space}
\end{center}
\vskip -0.2in
\end{figure}

\textbf{Figure \ref{observation-action-space}} (on the left) shows an example of Dead Or Alive++ observation where some of the \textit{RAM states} are highlighted, superimposed on the game frame.

An alternative configuration is also available, triggered by the \texttt{hardcore} setting: when set to \texttt{True} the observation space is composed only by the game frame, discarding all additional numerical values, thus it is of type \texttt{gym.spaces.Box}. Additional details can be found in the documentation.

\subsubsection{Action Space(s)}
\label{action-spaces-par}
Actions of the interfaced games can be grouped in two categories: move actions (up, left, etc.) and attack ones (punch, kick, etc.). 
DIAMBRA Arena provides four different action spaces: the main distinction is between \emph{discrete} and \emph{multi-discrete} ones. The former is a single list composed by the union of move and attack actions (of type \texttt{gym.spaces.Discrete}), while the latter consists of two sets combined, for move and attack actions respectively (of type \texttt{gym.spaces.MultiDiscrete}). 

For each of the two options, there is an additional differentiation available: if to use attack buttons combinations or not. This option is mainly available to reduce the action space size as much as possible, since combinations of attack buttons can be seen as additional attack buttons. 

The complete visual description of available action spaces is shown in Figure \ref{observation-action-space} (on the right), where all four choices are presented via the correspondent gamepad buttons configuration.

When run in 2P mode, the environment is provided with an action space of type  \texttt{gym.spaces.Dict} populated with two items, identified by keys \texttt{"P1"} and \texttt{"P2"}, whose values are either \texttt{gym.spaces.Discrete} or \texttt{gym.spaces.MultiDiscrete} as described above. Additional details on the action spaces are reported in the documentation.

\subsubsection{Reward Function}
\label{rew-funct-par}

The reward is defined as a function of characters' health values so that, qualitatively, damage suffered by the agent corresponds to a negative reward, and damage inflicted to the opponent corresponds to a positive reward. The quantitative, general, and formal reward function definition defined by \textbf{Equation \ref{reward-function}}:
\begin{equation}
R_t = \sum_i^{0,N_c}\left(\bar{H}_{i}^{t^-} - \bar{H}_{i}^{t} - \left(  \hat{H}_{i}^{t^-} - \hat{H}_{i}^{t}\right)\right)
\label{reward-function}
\end{equation}
where:
\begin{itemize}
\item $\bar{H}$ and $\hat{H}$ are health values for opponent's character(s) and agent's one(s) respectively; 
\item $t^-$ and $t$ are used to indicate conditions at "state-time" and at "new state-time" (that is before and after environment step); 
\item $N_c$ is the number of characters taking part in a round. Usually is $N_c = 1$ but there are some games where multiple characters are used, with the additional possible option of alternating them during gameplay, like Tekken Tag Tournament where 2 characters have to be selected and two opponents are faced every round (thus $N_c = 2$); 
\end{itemize}

\noindent The lower and upper bounds for the episode total cumulative reward are defined in \textbf{Equations \ref{min-max-reward}}. They consider the default reward function (Equation \ref{reward-function}) for game execution with \textit{continue game} option set equal to 0.0 (no continue allowed).

\begin{equation}
\begin{gathered}
\min{\sum_t^{0,T_s}R_t} = -N_c \left( \left(N_s-1\right) \left(N_r-1\right) +  N_r\right)\Delta H \\
\max{\sum_t^{0,T_s}R_t} = N_c N_s N_r\Delta H
\end{gathered}
\label{min-max-reward}
\end{equation}

where:
\begin{itemize}
\item $N_r$ is the number of rounds to win (or lose) in order to win (or lose) a stage; 
\item $T_s$ is the terminal state, reached when either $N_r$ rounds are lost (for both 1P and 2P mode) or game is cleared (for 1P mode only); 
\item $t$ represents the environment step and for an episode goes from 0 to $T_s$; 
\item $N_s$ is the maximum number of stages the agent can play before the game reaches $T_s$. 
\item $\Delta H = H_{max} - H_{min}$ is the difference between the maximum and minimum health values for the given game; usually, but not always, $H_{min}=0$.
\end{itemize}

For 1P mode $N_s$ is game-dependent, while for 2P mode $N_s=1$, meaning the episode always ends after a single stage (so after $N_r$ rounds have been won / lost be the same player, either P1 or P2).

For 2P mode, P1 reward is defined as $R$ in Equation \ref{reward-function} and P2 reward is equal to $-R$ (zero-sum games). Equation \ref{reward-function} describes the default reward function. It is of course possible to tweak it at will by means of custom \textit{reward wrappers} (see Section \ref{wrappers} below). Additional details can be found in the documentation.

\subsubsection{Wrappers}
\label{wrappers}
DIAMBRA Arena comes with a large number of ready-to-use wrappers and examples showing how to apply them. They cover a wide spectrum of use cases, and also provide reference templates to develop custom ones. In order to activate wrappers one has just to add an additional dictionary to the environment creation method, having properly populated it. A summary of available wrappers is presented in \textbf{Table \ref{wrappers-table}}, additional details can be found in the official documentation.

\begin{table*}[h]
\caption{Ready-to-use wrappers}
\begin{center}
\begin{small}
\begin{sc}
\begin{adjustbox}{max width=\textwidth}
\begin{tabular}{rl}
\toprule
Wrapper & Description \\
\midrule
Frame warping & \begin{tabular}{@{}l@{}}Resizes game frame to prescribed height and width, keeping it  RGB or making it grayscale \end{tabular}\\
\midrule
Observation scaling & \begin{tabular}{@{}l@{}}Scales elements of the observation space, with a specific operation per each observation type\end{tabular}\\
\midrule
\begin{tabular}{@{}r@{}}Frame stack with\\
  optional dilation \end{tabular} & 
\begin{tabular}{@{}l@{}}Stacks latest $N$ frames together piling them along the third dimension. Using the dilation factor, it allows \\
 to stack only one every $M$ frames, covering a broader temporal span for the same amount of memory\end{tabular}\\[0.2cm]
\midrule  
Actions stacking & 
Stacks latest $N$ actions together \\
\midrule  
\begin{tabular}{@{}r@{}}Observation dictionary\\
flattening and filtering \end{tabular} & 
\begin{tabular}{@{}l@{}}Flattens the observation dictionary nested levels and filters them keeping only the prescribed subset\end{tabular} \\
\midrule
Reward normalization &
\begin{tabular}{@{}l@{}}Scales the reward dividing it by the product of the scaling factor $K$ and $\Delta H$, difference between \\ maximum and minimum health, as previously described\end{tabular} \\[0.2cm]
\midrule
Reward clipping & Clips the reward applying the $sign\left(\right)$ function\\
\midrule
No-op reset & 
\begin{tabular}{@{}l@{}}Performs a maximum of $N$ no-op actions at the beginning of the episode (that is after reset)\end{tabular} \\
\midrule
Actions sticking &  
\begin{tabular}{@{}l@{}}Repeats the action sent to the environment for $N$ steps. It is activated when $N>1$ and can be applied\\
 only when \emph{Step Ratio} setting is equal to 1\end{tabular} \\[0.2cm]
\bottomrule
\end{tabular}
\end{adjustbox}
\end{sc}
\end{small}
\end{center}
\vskip -0.1in
\label{wrappers-table}
\end{table*}

\subsection{Advanced Features}

\subsubsection{Human-in-the-Loop Training}

Guiding the learning process by leveraging human input is desirable as taking advantage of humans domain expertise boosts learning, and helps the agent learning to behave as humans would expect. Different approaches have been studied by researchers \citep{Zhang19} such as humans providing evaluative feedback to the agent \citep{Knox08}, humans manipulating the agent’s observed states and actions \citep{Abel16}, or learning to imitate expert trajectories \citep{Ho16}. 

The success of the former two families of strategies strongly depends on how the human interfaces with the agent during learning, that is typically very difficult, if not impossible, with the majority of RL platforms and environments, since it requires low-level access to the environment mechanics that is rarely available.

DIAMBRA Arena can be easily set up to support an interactive and collaborative learning process between the human and the agent. The great flexibility given by environment wrappers, allows to easily include human contribution in the training process. It is possible for a human to have real-time access and interact with the agent during training in order to, for example, pause the scene, and control of the agent via keyboard or gamepad commands. 

Imitating expert trajectories is a significantly more common approach in research. DIAMBRA Arena provides advanced tools to leverage the \emph{imitation learning} technique \citep{Hussein17}. Very useful to speed up / bootstrap learning, usually in the very early training stages, it requires to store human gameplay in order to use it to train the agent policy, typically adopting either the \emph{supervised learning} approach (\emph{behavioural cloning}) or using the human to guide exploration in the "classic" RL setting. Recordings must therefore store both observations and actions sent by the human player at least in the former case, and rewards too in the latter.

The package comes with a wrapper dedicated to this purpose and one specific example showing how to set it up. In order to activate it, one has just to add an additional dictionary to the environment creation method.

Human players are requested to play using a USB gamepad, that is interfaced via a custom class named \texttt{DiambraGamepad}. Two main settings needs to be provided: \texttt{file\_path} and \texttt{user\_name}. The former is the local absolute path where to save recorded trajectories, provided as a string, while the latter is a string variable that can be used to differentiate between users in case recordings comes from multiple players.

\textit{Behavioral cloning approach.} Having a dataset (the collected player experience) with features (environment observations, for example the game frame) and targets (action(s) selected by the human player), one can train the agent policy network as typically done in \emph{supervised learning} classification problems. The result would be an agent whose behavior replicates the one of the player, that's where the term \emph{behavioral cloning} comes from.

\textit{Guided exploration approach.} Since rewards are stored alongside observations and actions, recorded trajectories can be used while remaining inside the RL paradigm: the human player has here the role of "guidance" in the exploration phase, providing the algorithm with a "meaningful" experience of the environment (from a human perspective) in the form of trajectories. This is expected to significantly speedup training, optimizing exploration towards zones of greater relevance in the observation/action domain space. 
\\

\lstset{frame=lines}
\lstset{label={imitationLearning}}
\lstset{basicstyle=\footnotesize}
\begin{lstlisting}[language=Python, caption=DIAMBRA Arena imitation learning example, captionpos=b]

import diambra.arena                                                                                                                            
                                                                                
settings = {"traj_files_list": ["path/to/recFile1",
                                "path/to/recFile2",
                                ...],
            "total_cpus": 2}
                                                                         
env = diambra.arena.ImitationLearning(**settings)                                                                
obs = env.reset()                                                       
                                                                                
while True:
    obs, rew, done, info = env.step(0)

    if done:
        obs = env.reset()
        break

env.close()
\end{lstlisting}

In order to do so, the software package provides a dedicated class named \texttt{ImitationLearning} with the specific purpose of loading trajectories recorded with the wrapper provided, and stepping through them. \textbf{Listing \ref{imitationLearning}} shows a code snippet with all basic instructions required to run it. 

It should be noted how in this case the action selection step is not needed anymore since the "history" of environment transitions is already written in stored human player experience files.

\subsubsection{Multi-Agent and Self-play}

As already mentioned, all environments can be run in both single player and two players mode, the latter making DIAMBRA Arena, de-facto, a software package that can be used for research in the fields of \emph{competitive multi-agent} and \emph{human-agent competition}.

This feature also allows to implement \emph{self-play} \citep{Baker20, Bansal18}, that consists, roughly speaking, in training an agent by making it play (or fight) against itself. In fact, supporting the 2P mode, the library allows the same agent (or two different agents) to play on both sides, one against the other.

This advanced technique offers many advantages: the algorithm always faces an opponent of similar skill level, enabling more efficient learning; it can be adopted also when no single player mode is available for a given game; it gives the agent the freedom to explore and find optimal strategies beyond those known by expert human players.

\subsection{Environments Comparison}
\label{sec-rlenv-comparison}

This section aims at comparing DIAMBRA Arena with the most important tools currently available in the literature that provide similar capabilities. Two main aspects are considered: features exposed by the software and its performances in terms of speed and memory footprint. The former is fundamental for measuring how broadly it can be applied, determining the range of research topics it enables to study. The latter is strictly linked with one of the most important limitations of current RL, sample complexity: the amount of data required by RL algorithms to achieve optimal performances is still very large, thus requiring very efficient environments that allow to generate the needed experience quickly and to perform a large number of experiments in reasonable time.

\subsubsection{Features}

\textbf{Table \ref{features-table}} compares different environments in terms of the most important characteristics for problems in this domain: type of observation spaces, type of action spaces, if they provide ready-to-use wrappers, and if they provide multi-agent and advanced features.

Regarding the observation spaces, DIAMBRA Arena covers all options provided by other tools but the full RAM state. This has been an explicit design choice, aiming to push the research towards studying algorithms able to learn in a more human-like condition, so avoiding the option of directly read the complete RAM states. Nonetheless, it can be easily added, if relevant, in future work.

In its current form, DIAMBRA Arena does not provide a continuous action space. This is mainly related to the nature of interfaced games, that are meant to be played with digital controllers (gamepads). Also in this case, extension to games featuring such type of input (for example mouse cursor position) will be subject of future work.

DIAMBRA Arena is one of the few environments providing ready-to-use wrappers. This is a key element in speeding up the implementation of an interface with third party RL libraries, such as Stable Baselines or Ray RLlib.

Lastly, a very relevant aspect is that DIAMBRA Arena allows to study many advanced topics, such as competitive agent-agent, human expert demonstration recording, imitation learning, transfer learning and cross-task generalization. In addition, it also natively covers human assisted reward and competitive human-agent settings, unique features not provided by the others, to the best knowledge of the author.

\begin{table*}[h]
\caption{Environments features comparison}
\begin{center}
\begin{small}
\begin{sc}
\begin{adjustbox}{max width=\textwidth}
\begin{tabular}{rccccc}
\toprule
Environment &
\begin{tabular}{@{}c@{}}Observation\\
 spaces \end{tabular} & 
\begin{tabular}{@{}c@{}}Action\\
 spaces \end{tabular} &
\begin{tabular}{@{}c@{}}Ready-to-use\\
 wrappers\end{tabular} &
 \begin{tabular}{@{}c@{}}Multi-agent\\
 ready \end{tabular} &
 \begin{tabular}{@{}c@{}}Advanced\\
 Features ready \end{tabular}\\
\midrule
  \begin{tabular}{@{}r@{}}DIAMBRA\\
 Arena \end{tabular}
& \begin{tabular}{@{}c@{}}Raw pixels (2D/3D),\\
Raw pixels (2D/3D) +\\ Vector data\end{tabular} & \begin{tabular}{@{}c@{}}Discrete,\\
Multi-discrete\end{tabular}& $\surd$ &   \begin{tabular}{@{}c@{}}Competitive agent-agent,\\
Competitive human-agent \end{tabular} & \begin{tabular}{@{}c@{}}Human expert \\ demonstration recording,\\
Imitation learning,\\ Human assisted reward, \\ Transfer learning, \\Cross-task generalization\end{tabular} \\[1.2cm]

  \begin{tabular}{@{}r@{}}Unity\\
 ML-Agents \end{tabular}
& Vector data & \begin{tabular}{@{}c@{}}Discrete,\\
Multi-discrete,\\Continuous\end{tabular}& X &  Competitive agent-agent & \begin{tabular}{@{}c@{}}Imitation learning,\\ Transfer learning,\\Cross-task generalization\end{tabular} \\[0.6cm]

StarCraft LE & \begin{tabular}{@{}c@{}}Raw Pixels (2D), \\Features layers,\\
Vector data\end{tabular} & Compound & $\surd$ & Competitive agent-agent & \begin{tabular}{@{}c@{}}Human expert \\demonstration recording\end{tabular} \\[0.6cm]

VizDoom & Raw pixels (3D) & Discrete & X &  Competitive agent-agent & \begin{tabular}{@{}c@{}}Human expert \\ demonstration recording\end{tabular} \\[0.4cm]

  \begin{tabular}{@{}r@{}}Unity\\
 Obstacle Tower \end{tabular}
& Raw pixels (2D/3D) & \begin{tabular}{@{}c@{}}Discrete,\\
Multi-discrete\end{tabular}& X &  - & \begin{tabular}{@{}c@{}}Transfer learning,\\Cross-task generalization\end{tabular} \\[0.6cm]

DMLab & \begin{tabular}{@{}c@{}}Raw pixels (3D), \\
Vector data\end{tabular} & \begin{tabular}{@{}c@{}}Discrete,\\
Continuous\end{tabular} & X & - & \begin{tabular}{@{}c@{}}Transfer learning,\\Cross-task generalization\end{tabular}  \\[0.4cm]

ALE & \begin{tabular}{@{}c@{}}Raw pixels (2D),\\
RAM state\end{tabular}& Discrete & X & - & - \\[0.4cm]

Gym & \begin{tabular}{@{}c@{}}Raw pixels (2D),\\
Vector data\\ RAM State\end{tabular} & \begin{tabular}{@{}c@{}}Discrete,\\
Multi-discrete,\\Continuous\end{tabular} & $\surd$ & - & - \\[0.6cm]

Retro & \begin{tabular}{@{}c@{}}Raw pixels (2D),\\
RAM state\end{tabular} & \begin{tabular}{@{}c@{}}Discrete,\\
Multi-discrete\end{tabular} & X & - & - \\[0.4cm]

MuJoCo & Vector data & Continuous & X & - & - \\[0.2cm]

\bottomrule
\end{tabular}
\end{adjustbox}
\end{sc}
\end{small}
\end{center}
\vskip -0.1in
\label{features-table}
\end{table*}

\subsubsection{Performances}

\textbf{Table \ref{speed-perf-table}} provides a comparison for the different environments in terms of execution speed, expressed in frames per second (FPS), and memory footprint for a single environment instance, measured in megabytes (MB). All environments have been run in the same machine, a standard low-medium level desktop also used to train DeepRL agents that is described in a later section.

Only most relevant environments are reported in the table, in particular those similar in terms of observation spaces, action spaces and advanced features. By looking at the values reported, one notes that both performance measures for DIAMBRA Arena are similar and around the same order of magnitude of the other RL environments. If compared with the most simple ones (ALE, VizDoom, Retro), the speed is between 2 to 6 times slower, while the memory footprint is between 1.5 to 4 times larger. This slightly lower efficiency is a fair price to pay to have the vast set of features DIAMBRA Arena provides that are not available in the other environments, as shown in the previous subsection. In fact, when compared with an environment of similar complexity (Unity Obstacle Tower), it becomes clear how faster DIAMBRA Arena is (about $10\times$), and how reasonable the memory footprint turns out to be. 

\begin{table}[!htb]
\caption{Environments performances comparison}
\begin{center}
\begin{small}
\begin{sc}
\begin{adjustbox}{max width=\textwidth}
\begin{tabular}{rcc}
\toprule
Environment &
Speed [FPS] & Memory [MB]\\
\midrule
  \begin{tabular}{@{}r@{}}DIAMBRA\\
 Arena \end{tabular}
& 0.5k & 140 \\[0.4cm]

  \begin{tabular}{@{}r@{}}Unity\\
 Obstacle Tower \end{tabular}
& 0.06k & 150 \\[0.4cm]

Retro & 1.1k & 86 \\[0.2cm]

DMLab & 1.2k & 40 \\[0.2cm]

VizDoom & 1.5k & 35 \\[0.2cm]

ALE & 3k & 70 \\[0.2cm]

\bottomrule
\end{tabular}
\end{adjustbox}
\end{sc}
\end{small}
\end{center}
\vskip -0.1in
\label{speed-perf-table}
\end{table}

\section{Trained DeepRL Agent}
\label{sec-agent}

In order to validate and confirm that the environments provided can be learned, many tests have been performed, training multiple DeepRL agents on different games, with different setups in terms of game settings, wrappers used, observation and actions spaces. 

This section describes in detail how these agents have been trained to play different games in single player mode, maximizing the total cumulative reward, where the immediate reward is defined by Equation \ref{reward-function}.

\subsection{Problem Framing and Algorithm}

To provide useful context for the discussion, in what follows the agent trained on Dead Or Alive++ will be considered. A reward normalization wrapper has been applied to it, using a scaling factor equal to $K=0.5$, the game has a $\Delta H = (H_{max}-H_{min}) = 208$, resulting in a normalization term equal to $K \Delta H = K (H_{max}-H_{min}) = 0.5*208 = 104$. It has 8 stages, the last always against the same final boss and the second last always against the same character that depends on the one used by the agent. Two rounds are needed to win a stage, and a single character is used. Therefore, with respect to quantities in Equations \ref{min-max-reward} and the normalization term defined above, one has $N_c = 1$, $N_s = 8$, $N_r = 2$, resulting in episode total cumulative reward bounds equal to $\min{\sum_t^{0,T_s}R_t} = -18$, $\max{\sum_t^{0,T_s}R_t} = 32$.

\begin{figure}[h]
\vskip 0.2in
\begin{center}
\centerline{\includegraphics[width=0.85\columnwidth]{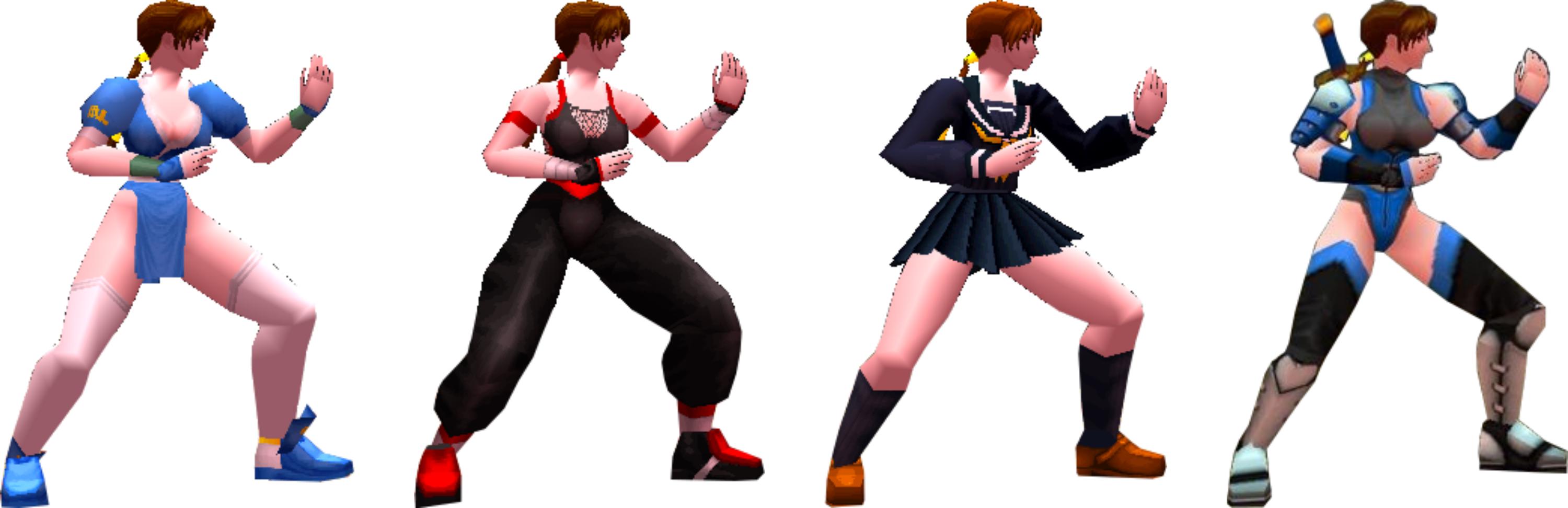}}
\caption{Kasumi available outfits in Dead Or Alive++}
\label{outfits}
\end{center}
\vskip -0.2in
\end{figure}

The selected game has four different difficulty levels, they do not affect other game settings. The native game frame resolution is $480\times 512\times 3$ and there are four different outfits for each character (see Figure \ref{outfits}).

Environment settings have been selected so that the agent learns to play with a specific character (Kasumi), as both P1 and P2 (it will randomly start the episodes in one of the two positions with equal probability), while using both the first two outfits for the selected character, sending actions once every 6 game steps (actions frequency 10 Hz, at fixed points in "time"). Settings are summarized in \textbf{Table \ref{environment-settings-table}}.

\begin{table}[!htb]
\caption{Environment settings used during training}
\begin{minipage}{0.5\textwidth}
\vskip 0.05in
\begin{center}
\begin{small}
\begin{sc}
\renewcommand{\arraystretch}{1.2}
\begin{tabular}{lc}
\toprule
Setting & Value \\
\midrule
Player & "Random" \\
Step Ratio & 6 \\
Frame Shape & [128,128,1] \\
Continue Game & 0.0 \\
Difficulty\footnote{Difficulty level has been set equal to 3 at the beginning of training, and has been raised to 4 when the agent was able to complete the game in the majority of evaluation episodes using the initial difficulty value.} & 3$\rightarrow$ 4 of 4 \\
Character & "Kasumi" \\
Character Outfit & 2 \\
Action Space & "Discrete" \\
Attack Buttons Combination & False \\
Hardcore & False \\
\bottomrule
\end{tabular}
\end{sc}
\end{small}
\end{center}
\vskip -0.1in
\label{environment-settings-table}
\end{minipage}
\end{table}

The selected observation space is made of the latest game frame plus the \emph{RAM states}. For training this agent only some elements of the latter have been selected, specifically: \emph{own/opponent health}, \emph{own/opponent side}, \emph{stage number} and \emph{actions}.

The \emph{discrete} action space was selected, with no attack buttons combination, for a total of 12 actions.

In addition to the previously mentioned \emph{reward normalization wrapper}, a number of additional ones have been applied: latest 4 frames are stacked together with no dilation; latest 12 actions are stacked together; observation scaling is applied; and no reward clipping, no-op reset nor actions sticking are applied. All settings are reported in \textbf{Table \ref{training-wrappers-table}}. 

\begin{table}[!htb]
\caption{Wrappers settings used during training}
\vskip 0.05in
\begin{center}
\begin{small}
\begin{sc}
\renewcommand{\arraystretch}{1.2}
\begin{tabular}{lc}
\toprule
Wrapper Settings & Value \\
\midrule
Observation Scaling & True \\
Frame Stacking with Dilation & $\left[4, 1\right]$ \\
Action Stacking & 12 \\
Reward Normalization & \begin{tabular}{@{}c@{}}True \\
(Scaling factor $K=0.5$) \end{tabular}\\[0.4cm]
Reward Clipping & False \\
No-Op Reset & 0 \\
Actions Sticking & 1 \\
\bottomrule
\end{tabular}
\end{sc}
\end{small}
\end{center}
\vskip -0.1in
\label{training-wrappers-table}
\end{table}

The proximal policy optimization (PPO) algorithm \citep{Schulman17} has been used, leveraging the open source RL library Stable Baselines \citep{StableBaselines}. The PPO2 model implemented therein has been interfaced with DIAMBRA Arena by means of a specific environment wrapper, mainly related to the management of observations format. A custom deep neural network has been designed for the policy and value networks, and provided to the PPO2 class as model to train. All details about networks model architecture, training strategy with hyperparameters and performances and results are described in the following three subsections.

\subsection{Model Architecture}

\begin{figure*}[h]
\vskip 0.2in
\begin{center}
\centerline{\includegraphics[width=0.9\textwidth]{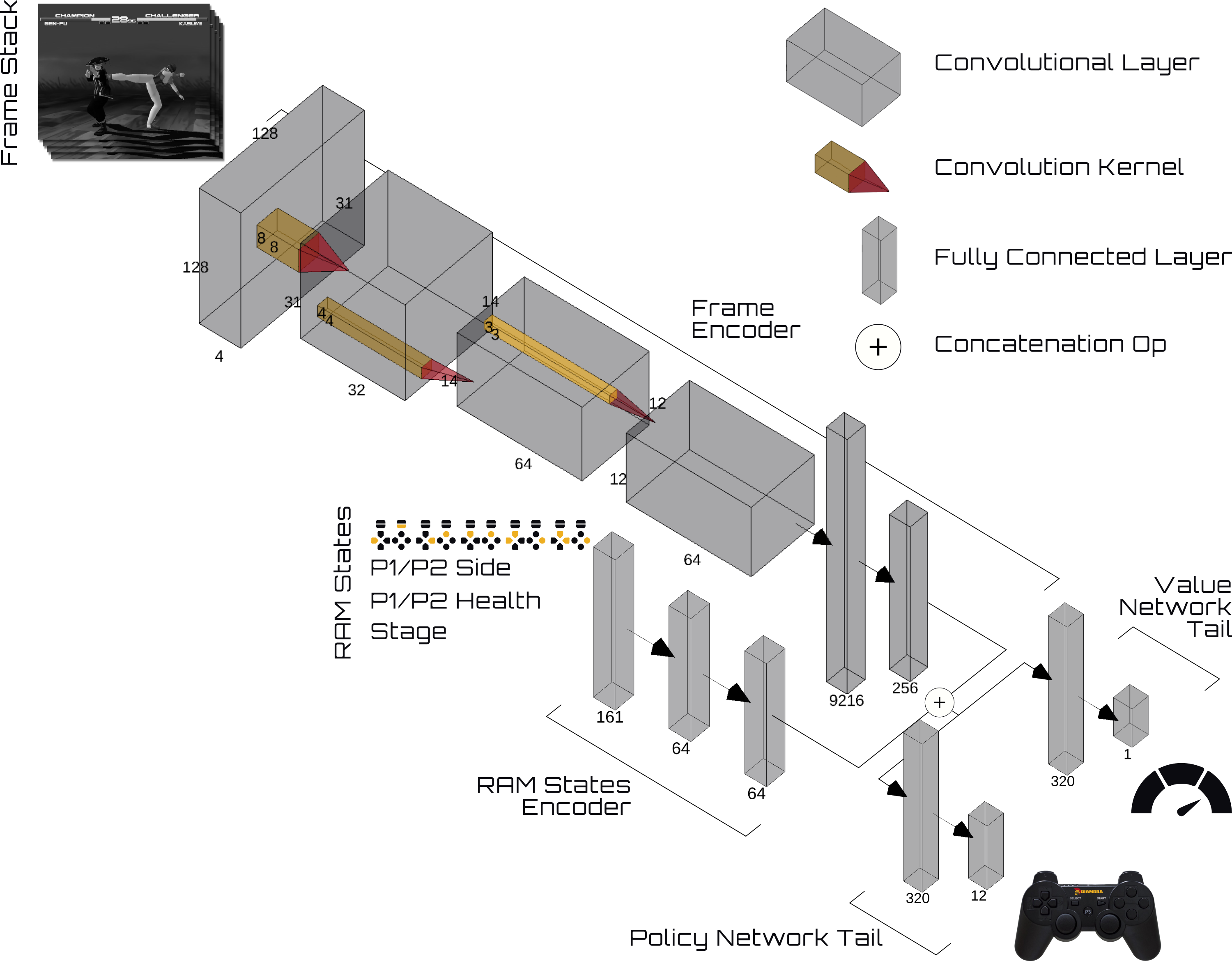}}
\caption{Model architecture scheme. Both the value function and the policy share the same backbone that extracts latent features from inputs, which are then forwarded to their specific layers.}
\label{model-architecture}
\end{center}
\vskip -0.2in
\end{figure*}

The \emph{policy} and the \emph{value networks} share all layers up to the latent space, where they bifurcate in two different tails. The shared part is composed by two different data processing pipelines, both performing information extraction, one from frames (\emph{frame encoder}) and the other one from the \emph{RAM states} (\emph{RAM states encoder}). The architecture of the two networks is described below, and also detailed in \textbf{Table \ref{network-architecture-table}} and in \textbf{Figure \ref{model-architecture}}. 

\begin{table}[!htb]
\caption{Model architecture details}
\vskip 0.05in
\begin{center}
\begin{small}
\begin{sc}
\renewcommand{\arraystretch}{1.2}
\begin{adjustbox}{max width=0.7\textwidth}
\begin{tabular}{lcl}
\toprule
Part & Layer Type & Details   \\
\midrule
Frame & In & $128\times 128\times 4$ tensor\\
Frame & Conv & $8\times 8$ kernel, 32 filters, \emph{relu}\\
Frame & Conv & $4\times 4$ kernel, 64 filters, \emph{relu}\\
Frame & Conv & $3\times 3$ kernel, 64 filters, \emph{relu}\\
Frame & FC/Out & 256 neurons (latent)\\
\midrule
RAM states & Input & $161\times 1\times 1$ tensor\\
RAM states & FC & 64 neurons, \emph{tanh}\\
RAM states & FC/Out & 64 neurons (latent), \emph{tanh}\\
\midrule
Concat & In & $256\times 1\times 1$ + $64\times 1\times 1$ tensor \\
Concat & Out & $320\times 1\times 1$ tensor\\
\midrule
Value & In & $320\times 1\times 1$ tensor\\
Value & Out & 1 value \\
\midrule
Policy & In & $320\times 1\times 1$ tensor\\
Policy & Out & 12 values (actions)\\
\bottomrule
\end{tabular}
\end{adjustbox}
\end{sc}
\end{small}
\end{center}
\vskip -0.1in
\label{network-architecture-table}
\end{table}

The frame encoder is almost the same as the one used in the Nature paper by \citet{Mnih2015}, usually referred as \emph{Nature CNN}. It is composed by three convolution layers plus a fully connected one; the input tensor is of size $128\times 128\times 4$ and a total of 256 latent features are generated by the final fully connected layer, which receives 9216 values obtained from flattening third convolution layer output tensor ($12\times 12\times 64$). It uses \emph{relu} as activation function.

The RAM states encoder is composed by two fully connected layers, generating 64 additional latent features. It uses \emph{tanh} as activation function.

Latent features extracted by the two encoders are concatenated together obtaining a 1D tensor of size $320$. 

The policy and value network tails are both made of a single fully connected layer connecting the $320$ latent features to, respectively: the set of $12$ available actions followed by a softmax block for action selection (policy net), and the single output value (value net).

\subsection{Training Strategy}

A fair amount of tests has been performed to identify a training strategy able to obtain good results. The resulting configuration consist of: $128\times 128$ frame resolution, grayscale frame depth, 4 frames stack, 12 actions stack, 16 parallel environments, 128 steps per update, batch size of 256, 4 training epochs per update, discount factor of 0.94, learning rate schedule $2e^{-4}\rightarrow 2e^{-6}$, clipping factor schedule $0.15\rightarrow 0.025$.

\begin{figure*}[h]
\vskip 0.2in
\begin{center}
\centerline{\includegraphics[width=0.9\textwidth]{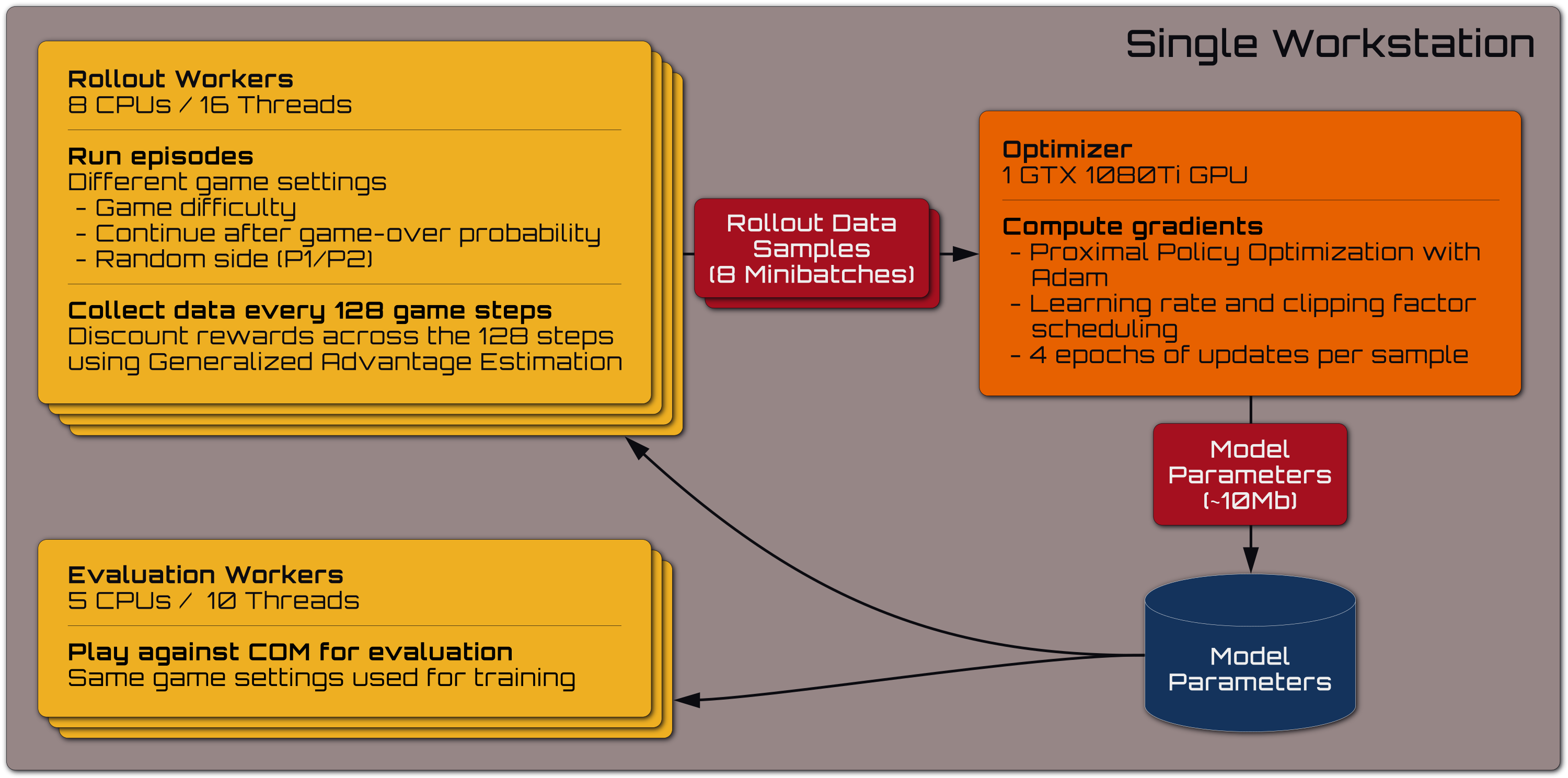}}
\caption{Block diagram representation of the training strategy}
\label{training-strategy}
\end{center}
\vskip -0.2in
\end{figure*}

Additional details are presented in \textbf{Table \ref{training-hyperparameters-table}} and \textbf{Figure \ref{training-strategy}}. They show the list of hyperparameters values with their description, and the training process block diagram. Three clusters of hyperparameters can be identified related to observation space, rollouts and training updates.

\begin{table*}[h]
\caption{Training hyperparameters}
\begin{center}
\begin{small}
\begin{sc}
\begin{adjustbox}{max width=\textwidth}
\begin{tabular}{lcl}
\toprule
Hyperparameter & Value & Details \\
\midrule
Frame size & $128\times 128$ & Game frame warping resolution\\[\tabVSpace]
Frame depth & 1 Channel & Game frame warping color compression \\[\tabVSpace]
Frame stack & $4$ & 
\begin{tabular}{@{}l@{}}Number of most recent frames experienced by the agent stacked \\
  together and used as input to the Policy and Value Networks\end{tabular}\\[0.4cm]
Actions stack & $12$ & 
\begin{tabular}{@{}l@{}}Number of most recent actions executed by the agent stacked together \\
and used in the RAM states as input to the Policy and Value Networks\end{tabular} \\[0.4cm]
\begin{tabular}{@{}l@{}}Parallel \\
 environments\end{tabular}
 & $16$ & 
Number of environments collecting training rollouts in parallel\\[0.4cm]
\begin{tabular}{@{}l@{}}Environment \\
 steps per update\end{tabular}
 & 128 & 
Number of environment steps run per update\\[0.4cm]
Batch size & $256$ & 
Number of training steps over which each Adam Optimizer update is computed\\[0.4cm]
\begin{tabular}{@{}l@{}}Training epochs\\
per update\end{tabular}
 & $4$ & 
Number of training epochs run per update\\[0.4cm]
Discount factor & $0.94$ & 
Discount factor used in return discounting, constant through training\\[0.4cm]
Learning rate & $2e^{-4}\rightarrow 2e^{-6}$ & 
\begin{tabular}{@{}l@{}}Learning rate used by Adam Optimizer update, linear scheduler function \\
 of training steps\end{tabular}\\[0.4cm]
Clipping factor & $0.15\rightarrow 0.025$ & 
\begin{tabular}{@{}l@{}}Clipping factor used by PPO surrogate objective clipping, linear scheduler \\
function of training steps\end{tabular}\\[0.4cm]
\bottomrule
\end{tabular}
\end{adjustbox}
\end{sc}
\end{small}
\end{center}
\vskip -0.1in
\label{training-hyperparameters-table}
\end{table*}

The first four hyperparameters (\emph{frame resolution}, \emph{frame depth}, \emph{frame stack} and \emph{actions stack}) have a direct impact on the size of the observation space. Different tests have been carried out, especially for \emph{frame resolution} and \emph{frame stack}, reaching values up to $\left[256\times 256\right]$ and $6$ respectively. The selected values, reported in Table \ref{training-hyperparameters-table}, allow to obtain a good performance while maintaining a reasonable training time, granting a $6\times$ speedup with respect to the largest configuration tested.

The next three hyperparameters (\emph{parallel environments}, \emph{environment steps per update} and \emph{batch size}) are strictly related to rollouts generation. They define the amount and the diversification of the experience collected by the agent, and also in this case, a trade-off is needed in order to limit RAM and GPU memory requirements. Having many environments running in parallel is particularly important to diversify the collected experience samples, minimizing their correlation, especially for standard on-policy methods (as PPO) that cannot leverage (optionally prioritized) memory replay buffers. The batch size plays a role too in this regard, while the number of environment steps per update must be chosen with care, taking into account in particular rewards sparsity, and making sure the value is high enough to collect informative experience samples, that is samples where collected rewards have relevant effects.

The last four hyperparameters (\emph{training epochs number}, \emph{discount factor}, \emph{learning rate} and \emph{clipping factor}) influence more low level aspects of training updates. Two key aspects to note of the adopted training strategy are: A) a linearly decreasing schedule, function of training steps, for both the learning rate and the clipping factor to favor fine convergence towards an optimal policy; B) the selected value for rewards discounting ($0.94$), which is again a parameter to be chosen with care. In fact, it determines how far back in time the rewards are propagated, thus it must be set taking into account rewards sparsity, past actions importance, and game specific dynamics.

The block scheme representing the complete training process (Figure \ref{training-strategy}) describes in detail the infrastructure: it consists of a single workstation in which 16 \emph{rollout workers} collect training experience in parallel, store experience collected in 128 environment steps and discount rewards across them. 10 \emph{evaluation workers} evaluate agent performances at regular intervals during training. Agent model parameters are shared between the two groups of workers. Rollout data samples are moved in batches to the single GPU to leverage CUDA computing, and model updates are performed accordingly to hyperparameters setup.

\subsection{Performances and Results}
\label{perf-res-sect}

\begin{figure}[h]
\vskip 0.2in
\begin{center}
\centerline{\includegraphics[width=0.75\columnwidth]{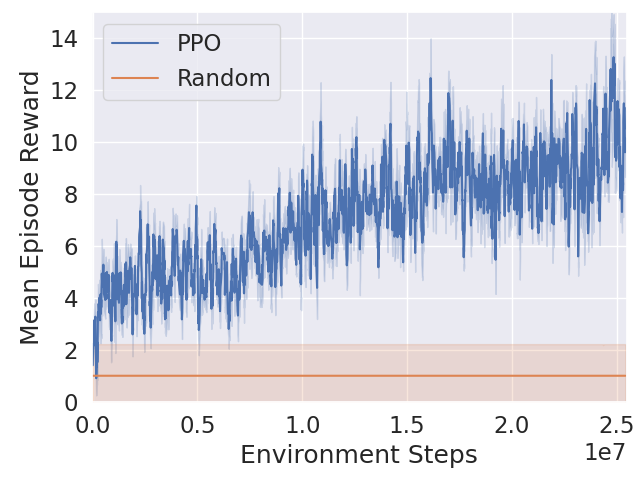}}
\caption{Mean episode reward plot as a function of environment training steps, with random agent score as a reference baseline}
\label{training-reward}
\end{center}
\vskip -0.2in
\end{figure}

Two different hardware setups have been used to perform training tests, \textbf{Table \ref{specs-numbers-table}} provides details in terms of components and training speed. They can be considered low-to-medium level home desktop configurations in terms of computing power. Agent performance as a function of training steps is presented in \textbf{Figure \ref{training-reward}}, where the 10-episodes averaged total cumulative reward for the PPO agent is plotted together with the random agent one. After 25M training steps, the trained agent average reward has reached a value of around 12, notably larger than the random agent one, a clear demonstration it is learning how to play in order to maximize episode total cumulative reward. Visual comparison between the random agent and the trained one demonstrates, even more than the chart, how well the agent policy learned a playing strategy at least very good, if not optimal. 

\begin{table}[!htb]
\caption{Hardware specs and training numbers}
\vskip 0.05in
\begin{center}
\begin{small}
\begin{sc}
\begin{adjustbox}{max width=0.85\textwidth}
\begin{tabular}{lcc}
\toprule
& Setup \#1 & Setup \#2 \\
\midrule
CPU &  \begin{tabular}{@{}c@{}}i5 4 cores/  \\
       8 threads\end{tabular} &
       \begin{tabular}{@{}c@{}}AMD Rayzen 9  \\
        12 cores/24 threads\end{tabular}       \\[\tabVSpaceDouble]
RAM & 16Gb & 16Gb\\[\tabVSpace]
GPU & GTX 1050 4Gb & GTX 1080Ti 11Gb\\[\tabVSpace]
\midrule
Env steps / day & $\mathtt{\sim}$2.6M & $\mathtt{\sim}$10.2M \\
Training time & $\mathtt{\sim}$10.2 Days 24/7 & $\mathtt{\sim}$2.36 Days 24/7\\
\bottomrule
\end{tabular}
\end{adjustbox}
\end{sc}
\end{small}
\end{center}
\vskip -0.1in
\label{specs-numbers-table}
\end{table}

A video comparison showing the agent at three different training stages (10M steps, 25M steps and 50M steps) can be found at \href{https://www.youtube.com/watch?v=2IXsMAdAEBU}{this link}\footnote{DeepRL Agent in Dead Or Alive: \url{https://www.youtube.com/watch?v=2IXsMAdAEBU}}. There are a few elements clearly showing agent improvement, two that can be easily noted are: A) the trained agent stops performing attack moves when the opponent is far and out of reach. Instead, it starts to wait for the opponent not only to be close enough, but even to stand up when laying on the ground after being hit; B) the trained agent timely performs counter-moves when appropriate, evading opponent's attacks.

An identical approach has been successfully adopted to train the same DeepRL agent in the other games and omitted here for conciseness. Visual results are publicly available for Street Fighter III and Tekken Tag Tournament at \href{https://www.youtube.com/watch?v=dw72POyqcqk}{this link}\footnote{DeepRL Agent in Street Fighter: \url{https://www.youtube.com/watch?v=dw72POyqcqk}} and \href{https://www.youtube.com/watch?v=XEJ9QfmmzwM}{this link}\footnote{DeepRL Agent in Tekken Tag \url{https://www.youtube.com/watch?v=XEJ9QfmmzwM}} respectively.

Achieving these results within a training time of less than two and a half days, using a medium-low level standard home desktop, confirms these environments, while featuring games with complex mechanics, can be successfully used by everyone.

\section{Discussion}
\label{sec-discussion}

The presentation of DIAMBRA Arena (Section \ref{sec-arena}) and DeepRL Agent training results (Section \ref{sec-agent}), provides many interesting insights. One important advantage of the software package is the high degree of customization in terms of action spaces, observation space and environments mechanics. For example, it is possible to make the tasks purely visual (using the screen buffer alone as observation) or to consider a selection of (fair) RAM values too as available features, to easily extend the actions set or to use multiple characters.  

The software comes with ready-to-use wrappers of different kind, covering observation, action and reward wrapping classes. These are very important elements when it comes to interfacing environments with RL libraries, taking advantage of already implemented ones results in a major speedup.

DIAMBRA Arena guarantees environments accessibility and usability for a very broad audience. Thanks to minimal requirements in terms of computing power, even CPU-only training is a viable option. The comparison with similar tools currently available in the literature, carried out in Section \ref{sec-rlenv-comparison}, demonstrated a comparable performance in terms of speed and memory footprint, while providing more advanced environments enabling research on more advanced and complex RL topics. In addition, Section \ref{sec-agent} showed only a few days are needed to train end-to-end a DeepRL agent with a low-medium level workstation. It can be easily interfaced with third party libraries, especially for RL training, as done in this work with Stable Baselines.

In its current form, DIAMBRA Arena provides fighting games as RL tasks, covering many different interesting options for modern RL research that are discussed below. Moreover, additional types of environments are already planned for the near future.

\textit{Agent generalization.} One potential problem of the majority of RL benchmarks is that they are not ideal for testing generalization between similar tasks. This may cause falling into the pitfall of \textit{"training RL algorithms on the test set"}, measuring model performance on the same environment(s) it was trained on. In order to prevent this, RL environments need to provide means by which they can replicate the concept of "train" and "test" split, as typically done for supervised learning datasets. Few-shot learning, cross-task generalization, exploration-maximization, are all topics that would relevantly benefit from such a feature. DIAMBRA Arena already supports this need, providing different tasks with similar scope and mechanics; it allows to run episodes from game start to end, as a sequence of different stages; and allows to select different game difficulty levels. 

These are all features that can be used to test agent generalization capabilities. To stress this aspect even more, the future addition of new games will favor environments consisting of many similar tasks sampled from a single task distribution creating opportunities to learn how to explore on some levels and transfer this ability to other levels.

\textit{Curriculum and transfer learning.} Typically, RL challenges and tasks are approached with the aim of solving them from scratch. One of the most interesting directions of current research considers sequences of tasks, training the algorithm on one task after the other. These sequences are made of tasks with an increasing level of difficulty, and are meant to be solved in order. As for agent generalization, also these applications are already supported and more effort will be put to extend features in this specific direction.

\textit{Human-in-the-loop training.} Functionalities currently available in the software package allow to easily provide feedback to the agent and to modify the environment during training, thus supporting exploration in this field of research. Also in this case, new types of environments are planned for development in the near future, with the aim of making human-in-the-loop training even more accessible for research and experimentation.

\textit{Multi-agent.} Current version of DIAMBRA Arena already supports multi-agent applications in the competitive flavor (both agent-agent and agent-human). One of the future development directions is to add new environments in which also the cooperative flavor is available, for both agent-agent as well as human-agent settings. Additional directions are being discussed, also touching the so called "value alignment problem", one of the aspects involved in typical reasonings around existential concerns for AI.

\textit{Real-world operation.} The need to validate RL algorithms in the real world is of paramount importance, and would represent a major achievement that could lead to the realization of a great potential. Almost certainly, it will involve aspects related to agent-human cooperation/competition, thus requiring robust and scalable training where agents play against (or in cooperation with) humans. The aim of developing tools and infrastructure to enable the study of agent-human interaction at scale is also in the road map.

\section{Conclusions}
\label{sec-conclusions}

This paper presented DIAMBRA Arena, a novel software package for RL research and experimentation. It provides high-quality environments, complemented by a very broad set of tools enabling many different lines of research that focus on major challenges the scientific community is facing, as \emph{standard RL}, \emph{competitive multi-agent}, \emph{human-agent competition}, \emph{human-in-the-loop training}, \emph{imitation learning} and \emph{self-play}. Supporting all major OSs, and implementing the standard OpenAI Gym Python API interface, its adoption is easy and straightforward. Multiple DeepRL agents have been trained in the provided environments, using PPO algorithm through Stable Baselines library, obtaining optimal performances and human-like behavior. Results achieved confirm DIAMBRA Arena utility as a reinforcement learning research platform, providing environments designed to study the most challenging topics in the field.

\section*{Acknowledgments}

The author thanks Marco Meloni for his insightful comments and suggestions for editing an earlier draft of this manuscript.






\newpage
  \bibliographystyle{elsarticle-harv} 
  \bibliography{diambra-arena}


%
%
%
\end{document}